\documentclass[twocolumn,final]{elsarticle}
\usepackage[colorlinks=true,linkcolor=blue]{hyperref}
\usepackage{lineno}
\usepackage{multirow}
\usepackage{soul}
\modulolinenumbers[1]

\biboptions{sort&compress}

\begin{document}
\begin{frontmatter}
\title{{\color{blue}Knowledge discovery from emergency ambulance dispatch during COVID-19: A case study of Nagoya City, Japan}}
\author[a,b]{Essam A. Rashed}
\ead{essam.rashed@nitech.ac.jp}
\author[a]{Sachiko Kodera}
\author[c]{Hidenobu Shirakami}
\author[c]{Ryotetsu Kawaguchi}
\author[c]{Kazuhiro Watanabe}
\author[a,d,e]{Akimasa Hirata}
\address[a]{Department of Electrical and Mechanical Engineering, Nagoya Institute of Technology, Nagoya 466-8555, Japan}
\address[b]{Department of Mathematics, Faculty of Science, Suez Canal University, Ismailia 41522, Egypt}
\address[c]{Nagoya City Fire Department, Nagoya, Aichi, Japan}
\address[d]{Center of Biomedical Physics and Information Technology, Nagoya Institute of Technology, Nagoya 466-8555, Japan}
\address[e]{Frontier Research Institute for Information Science, Nagoya Institute of Technology, Nagoya 466-8555, Japan}


\begin{abstract}

Accurate forecasting of medical service requirements is an important big data problem that is crucial for resource management in critical times such as natural disasters and pandemics. With the global spread of coronavirus disease 2019 (COVID-19), several concerns have been raised regarding the ability of medical systems to handle sudden changes in the daily routines of healthcare providers. One significant problem is the management of ambulance dispatch and control during a pandemic. To help address this problem, we first analyze ambulance dispatch data records from April 2014 to August 2020 for Nagoya City, Japan. Significant changes were observed in the data during the pandemic, including the state of emergency (SoE) declared across Japan. In this study, we propose a deep learning framework based on recurrent neural networks to estimate the number of emergency ambulance dispatches (EADs) during a SoE. The fusion of data includes environmental factors, the localization data of mobile phone users, and the past history of EADs, thereby providing a general framework for knowledge discovery and better resource management. The results indicate that the proposed blend of training data can be used efficiently in a real-world estimation of EAD requirements during periods of high uncertainties such as pandemics.

\end{abstract}


\end{frontmatter}



\section{Introduction}

An ambulance is one of the most important healthcare tools providing an essential life-saving role on a daily basis. The management and location of the ambulance dispatch center are known to reduce the rate of fatalities, particularly during a national emergency caused by a natural disaster or wide-spread pandemic. The outbreak of the infectious coronavirus disease 2019 (COVID-19) was reported in China in 2019~\cite{Wu2020JAMA}, and the disease spread throughout the globe during the early months of 2020. Owing to its fast and significant spread, a recognized collapse in medical systems and shortages of medical equipment have been reported in several countries ~\cite{Armocida2020TLPH, Ranney2020NEJM,Raoofi2020AIM}. Clinical symptoms of COVID-19 include fever, cough, and difficulty breathing, which are not radically different from those of other seasonal infections in their early stage~\cite{Sohrabi2020IJS}, which increases the challenge for the accurate diagnosis and treatment. To prevent virus infection at hospitals, an early pandemic protocol was considered~\cite{Glauser2020CMAJ}. Emergency service management should be included in this protocol. Later on, a guidance for health worker infection prevesion was relased by World Health Organization (WHO)~\cite{WHO2020}. Paramedics have also suffered from the spread of this novel virus; thus, a more careful management than usual is required~\cite{Whitfield2020AJP, Buick2020CJEM,Higginson2020JPP}. 

In Japan, the first state of emergency (SoE) was declared nationwide on April 16, 2020, and was revoked on May 25, 2020. Note that, during this pandemic, a complete closure policy was not adopted in Japan, but rather a voluntary segregation and seclusion with community cooperation was applied. During the SoE, outdoor activities were reduced, particularly in common crowded regions such as major train stations. From April 18, 2020, NTT Docomo, Inc. (a mobile phone operator in Japan) started to provide publicly available data on activities based on the user locations of mobile phones or smartphones in major traffic stations nationwide.

The careful management of healthcare infrastructure is required for fair allocation and usage especially where these resources are limited~\cite{Emanuel2020NEJM}. Owing to a lack of data on ambulance services in such a pandemic era, it was unclear how many allocations were needed for ambulances, including the number of emergency ambulance dispatches (EADs) and special requirements for additional processes, such as disinfection after the transport of potentially positive cases. The open questions here are i) how accurate EAD forecasts can be when applying environmental factors, ii) how the changes caused by abnormalities such as a SoE during a pandemic should be handled, and iii) what are the main reasons for such changes, including the relationship with human activities? If such information is made available, additional measures for ambulance management can be implemented. For example, it will be possible to strictly limit the use of certain ambulance units for potential COVID-19 patients. 

\begin{figure}
\centering
\includegraphics[width=.4\textwidth]{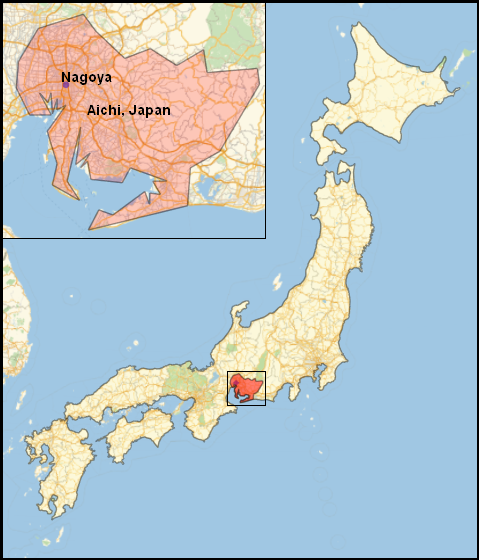}
\caption{Map of Japan with Aichi prefecture and Nagoya City highlighted.}
\label{map}
\end{figure}

This study first analyzed EAD data recorded from April 1, 2014, to August 18, 2020, in Nagoya City, Japan. This analysis has led to a high-quality estimation of the number of EADs based on data gathered before the COVID-19 era using machine learning approaches. A comparison of the estimated and actual EADs observed during the pandemic clarifies the differences caused by the COVID-19 outbreak. A data analysis model can provide better understanding of the potential approaches used to estimate the number of EADs during a pandemic and calls during a SoE. The main factor for this was also discussed for potential future pandemics, including third waves of COVID-19. To the best of the authors’ knowledge, this is the first study that applies a deep learning approach to forecast the daily number of EADs when considering environmental factors, even during non-pandemic states. The main contributions of this study can be summarized as follows:

\begin{itemize}
\item A machine learning architecture for accurate EAD forecasting in urban areas from environmental factors
\item A trained long short-term memory (LSTM) network for EAD estimation in Nagoya City, Japan, with a potential extension to other regions based on data availability
\item The introduction of a new social factor (i.e., mobile phone usage) that can be used to fine-tune the EAD forecasting during a pandemic
\end{itemize}


\section{Materials and methods}

\begin{figure*}
\centering
\includegraphics[width=\textwidth]{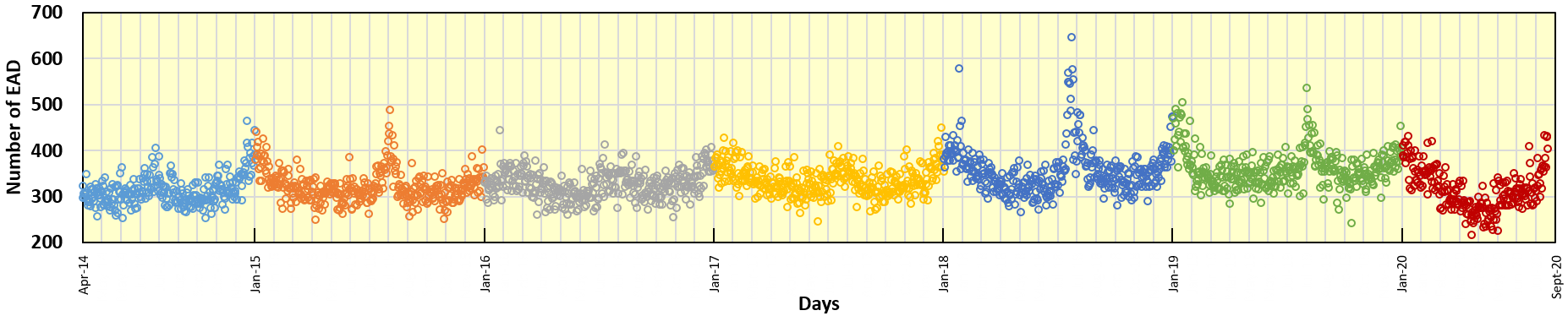}
\caption{Daily total number of recorded EADs in Nagoya City from April 1, 2014 to August 19, 2020. Years are presented in different colors.}
\label{all_EAD_all_time}
\end{figure*}
\subsection{Study area and data source}

Nagoya is a major city in Japan located in the central region of Honshu Island (Fig.~\ref{map}) with a population varied from 2,272$k$ to 2,316$k$ (2014--2020), which makes Nagoya the fourth largest city nationwide\footnote{\href{http://www.city.nagoya.jp/shisei/category/67-5-5-5-0-0-0-0-0-0.html}{http://www.city.nagoya.jp (21 Aug. 2020)}}. All daily EAD data were collected by authorities at the Fire Department of the City. Fig.~\ref{all_EAD_all_time} illustrates the daily total number of EADs from April 2014 to August 2020. The total number of EADs can be thought of as a U-shaped plot with two annual peaks during the summer and winter, which is highly associated with upper and/or lower temperature peaks. Weather data, including maximum and/or minimum daily temperature and other related factors such as humidity, were collected for Nagoya City from the online resources of the Japan Meteorological Agency\footnote{\href{https://www.jma.go.jp/jma/indexe.html}{https://www.jma.go.jp/jma/indexe.html (21 Aug. 2020)}}. We also processed data representing the variations (in percentage) of people around the major transport stations in Japan collected by the mobile career NTT Docomo, Inc. and released on the web\footnote{\href{https://mobaku.jp/covid-19/}{https://mobaku.jp/covid-19/ (21 Aug. 2020)}}\textsuperscript{,}\footnote{\href{https://www.nttdocomo.co.jp/utility/demographic\_analytics/}{https://www.nttdocomo.co.jp (21 Aug. 2020)}}. These data were made available immediately after the emergency declaration on April 18, 2020. Note that the national market share of NTT Docomo is approximately 36.9\% (ranked first)\footnote{\href{https://www.soumu.go.jp/main\_sosiki/joho\_tsusin/eng/pressrelease/2020/12/18\_01.html}{https://www.soumu.go.jp (2 Feb. 2021)}}. The data are based on the estimated statistical population generated from mobile terminal network operational data~\cite{Terada2013NTT}.

\begin{figure}
\centering
\includegraphics[width=0.5\textwidth]{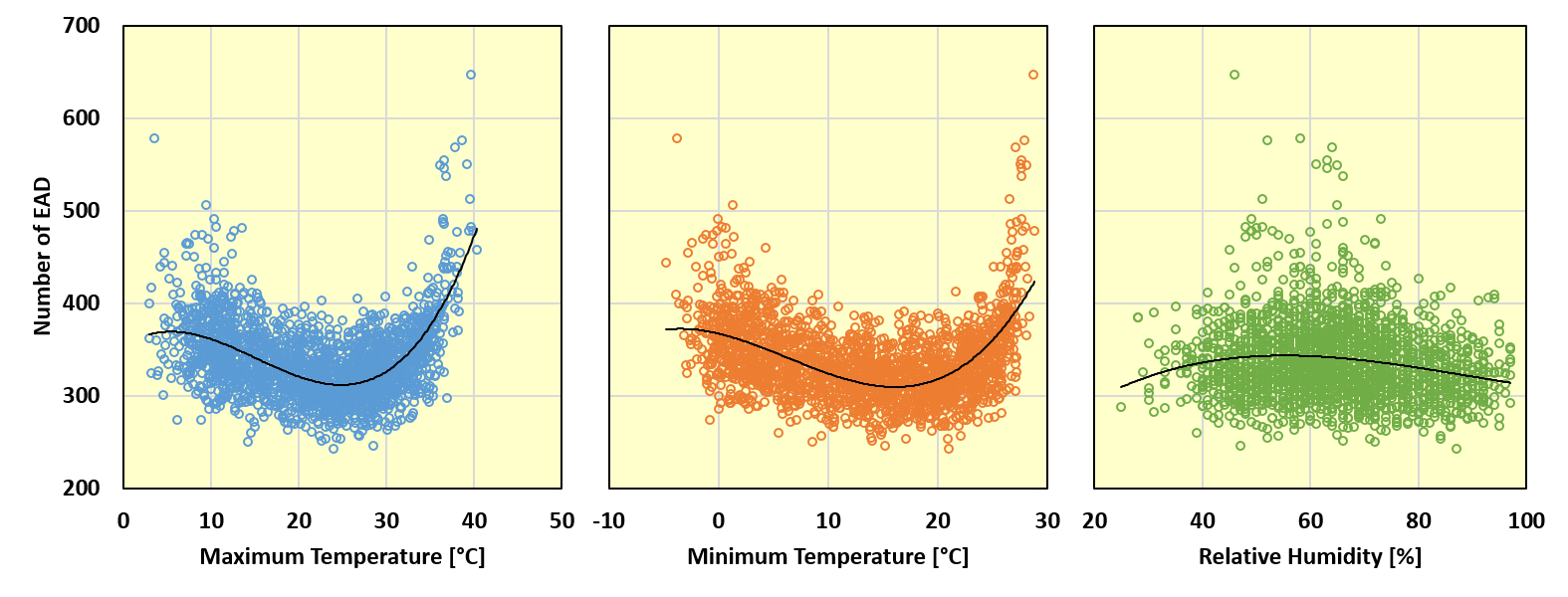}
\caption{Correlation between the daily total EAD and the maximum (left) and minimum (middle) temperatures and relative humidity (right) using data from April 1, 2014 to December 31, 2019.}
\label{tmphum}
\end{figure}


\subsection{Correlation with environmental factors}

It is well known that the number of ambulance dispatches is related to the daily average ambient temperature~\cite{Alessandrini2011ER, Bassil2011JECH, Cheng2016IJB, Kotani2018GHA, Sangkharat2020ER, Patel2019EP, Hu2020EP}. The effects of the daily maximum and minimum ambient temperatures, as well as the relative humidity, on the number of daily EADs are shown in Fig.~\ref{tmphum}. A common U-shape curve can be observed at both the maximum and minimum temperatures, whereas the relative humidity triggers a slightly higher number of EADs within the middle range. However, the effect of environmental factors can be split into different patterns based on the cause or illness corresponding to the EADs.

In addition, weekends and holidays are other factors that characterize human social activities. Furthermore, the age of the population affects such activities. The typical retirement age in Japan is currently approximately 60-65 years. In addition, an age of over 65 is classified as elderly; thus, 65 years is used as a reference value in this study. The former factors were used as input for applying machine learning to the data on the previous 5 years. The statistics are applied for the populations younger and older than 65 years in age. The learning-based estimation is applied as a reference value for 2020 and then compared with the observed value determined by the city fire department.
\begin{figure*}
\centering
\includegraphics[width=\textwidth]{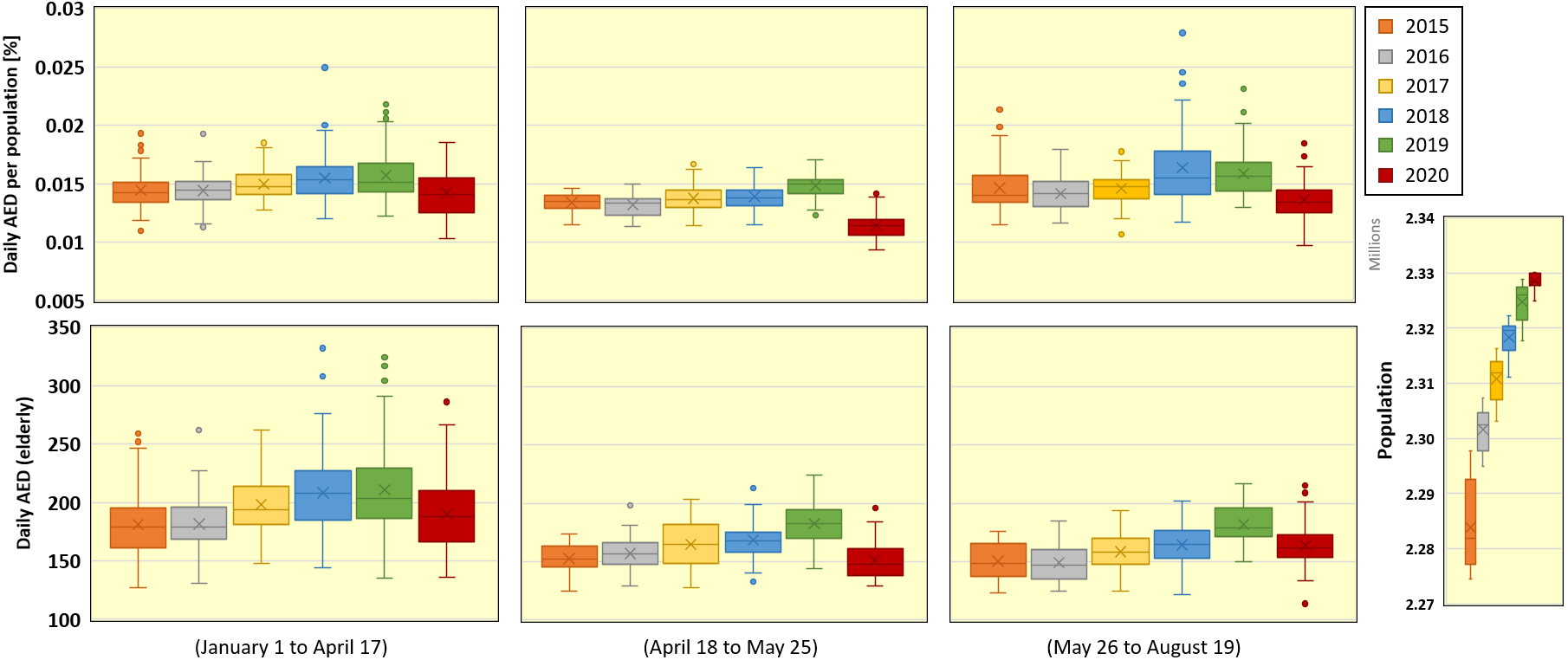}
\caption{Box plot of daily EAD calls during 2020 (January 1 to April 17, April 18 to May 25, and May 26 to August 19) compared with the same time periods from 2015 to 2019. Top row is the percentage of daily AED per population and bottom row is the AED numbers for elderly (age greater than 65 years) citizens. These time periods are before, during, and after the first SoE declaration in Japan to mitigate COVID-19 pandemic. Right-side graph demonstrate the Nagoya city population.}
\label{avgEADallTime}
\end{figure*}
\subsection{Effects of the Covid-19 pandemic}

\begin{figure}
\centering
\includegraphics[width=0.5\textwidth]{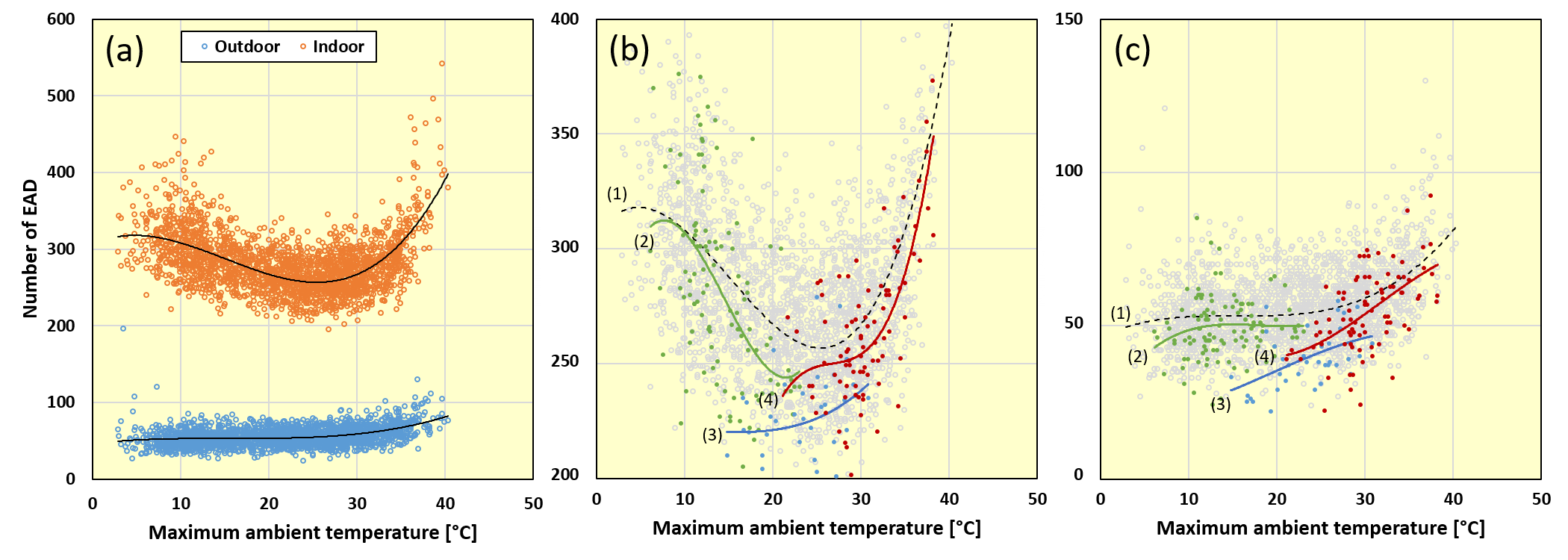}
\caption{Ambulance dispatches in Nagoya City are associated with the daily maximum ambient temperature for patients selected based on their indoor and outdoor activities. (a) Plots of all data measured during a normal state (April 2014 to December 2019). (b) and (c) plots of both normal (gray) and after outbreak for indoor and outdoor calls, respectively. Data and their regression curve with label (1) refer to (a). Regression curves with labels (2), (3), and (4) indicate the start of the COVID-19 pandemic spread but prior to the first SoE declaration (January 1, 2020, to April 17, 2020), during the first SoE (April 18 to 25 May) and after the first SoE (May 26 to August 19), respectively. The vertical scale is magnified.}
\label{ADplace}
\end{figure}

Unexpected pandemics are known to cause extensive requirements for medical care services and special resource management~\cite{Bielajs2008PDM}. From the data presented in Fig.~\ref{avgEADallTime}, clearly, the number of EADs in 2020 is lower than that during the past 5 years. Figure~\ref{ADplace} shows the number of EADs according to the maximum daily ambient temperature. Data were divided into categories based on the pickup site (i.e., indoor or outdoor). The normal pattern observed since April 2014 significantly changed during the pandemic, as smaller numbers of dispatches were observed in both location categories. A further reduction was also recognized during and after the first SoE (up to August 19). One potential reason for the reduction in EADs during the pandemic is the change in social activities. However, there are several other factors such as the enhanced hygiene~\cite{Cerulli2020JNS} and avoiding the access to medical facilities to prevent potential COVID-19 infection~\cite{ lange2020}. It is difficult to explicitly highlight the main reason but it is likely a composition of many factors include the above mentioned ones. Even using models that enable artificial intelligence (or machine learning), specific data patterns are challenging to estimate using models trained on data measured under normal situations.

\subsection{Recurrent Neural Network}

\begin{figure*}
\centering
\includegraphics[width=\textwidth]{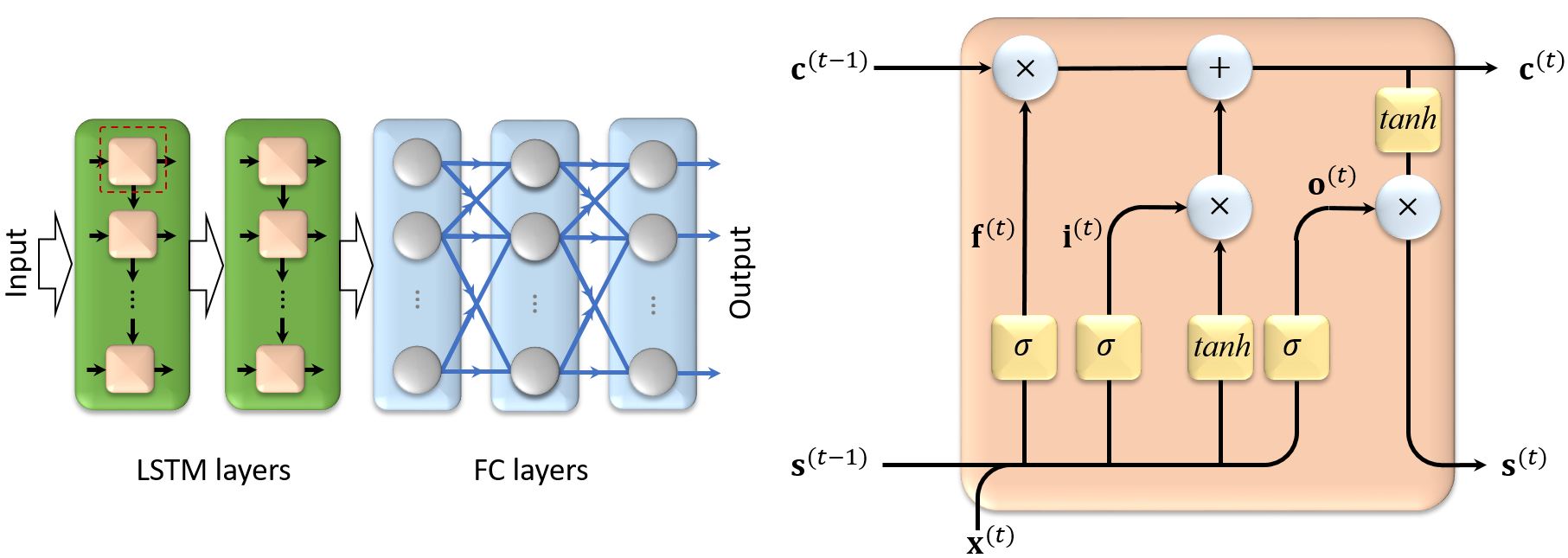}
\caption{Left is the network architecture and right is the basic structure of a single LSTM node.}
\label{lstm}
\end{figure*}

Recurrent neural networks (RNNs) are a wide range of network architectures that consider inter-neuronal connections such that they formulate a memory-like unit~\cite{Graves2009TPAMI}. LSTM is a commonly used class of RNN that performs well with long-term dependencies (Fig.~\ref{lstm}). To address this problem, we consider an LSTM-based network architecture. Consider a time-series data sample ${\bf X}=({\bf x}^{(1)},{\bf x}^{(2)}, \dots, {\bf x}^{(T)})^{\top}$, where ${\bf x}^{(t)}=(x_1^{(t)},x_2^{(t)},\dots,x_{I}^{(t)})$ and $x_i^{(t)}$ is the measurement of the $i$th variable at the $t$th time slot frame. The computation within a single LSTM node can be expressed as follows:

\begin{equation}
{\bf i}^{(t)}=\sigma({\bf W}_{ix} {\bf x}^{(t)} + {\bf W}_{is} {\bf s}^{(t-1)} + {\bf b}_i), 
\end{equation}
\begin{equation}
{\bf o}^{(t)}=\sigma({\bf W}_{ox} {\bf x}^{(t)} + {\bf W}_{os} {\bf s}^{(t-1)} + {\bf b}_o), 
\end{equation}
\begin{equation}
{\bf f}^{(t)}=\sigma({\bf W}_{fx} {\bf x}^{(t)} + {\bf W}_{fs} {\bf s}^{(t-1)} + {\bf b}_f), 
\end{equation}
\begin{equation}
{\bf m}^{(t)}=\phi({\bf W}_{mx} {\bf x}^{(t)} + {\bf W}_{hc} {\bf s}^{(t-1)} + {\bf b}_m), 
\end{equation}
\begin{equation}
{\bf c}^{(t)}={\bf f}^{t} \odot {\bf c}^{(t-1)} + {\bf i}^{t} \odot {\bf m}^{(t-1)}, 
\end{equation}
\begin{equation}
{\bf s}^{(t)}={\bf o}^{(t)} \odot \phi({\bf c}^{(t)}),
\end{equation}
where ${\bf i}^{(t)}$, ${\bf o}^{(t)}$, ${\bf f}^{(t)}$, ${\bf m}^{(t)}$, and ${\bf s}^{(t)}$ represent the input gate, output gate, forget gate, memory gate, and node state values at time frame $t$ (here, day index), respectively. ${\bf W}$ and ${\bf b}$ are the node weights (parameters) and bias matrices. $\sigma$ and $\phi$ are the sigmoid and $tanh$ functions, respectively. LSTM layers are trained using time-series data to optimize the parameters for future forecasting tasks.

The proposed architecture consists of two LSTM layers (with 50 and 30 nodes) and a three fully connected (FC) layers (with 300, 100 and $K$ nodes). The input data vector contains the daily based maximum ambient temperature, average relative humidity, and a binary label that identifies working days from non-working days (weekends or national holidays). The network output is a $K$ nodes representing the number future days to be predicted. The network parameters (${\bf W}$) and bias (${\bf b}$) are initialized with zero matrices. In addition, the Adam algorithm~\cite{Kingma2014arXiv} is used for data fitting during training along with the cross-entropy loss function and automatically estimated learning rate. The network architecture was implemented using Wolfram Mathematica \textregistered~ver. 12.1 on a workstation with four Intel \textregistered~Xeon CPUs running at 3.60 GHz, with 128 GB of memory and three NIVIDIA GeForce 1080 GPUs. 

\begin{figure*}
\centering
\includegraphics[width=\textwidth]{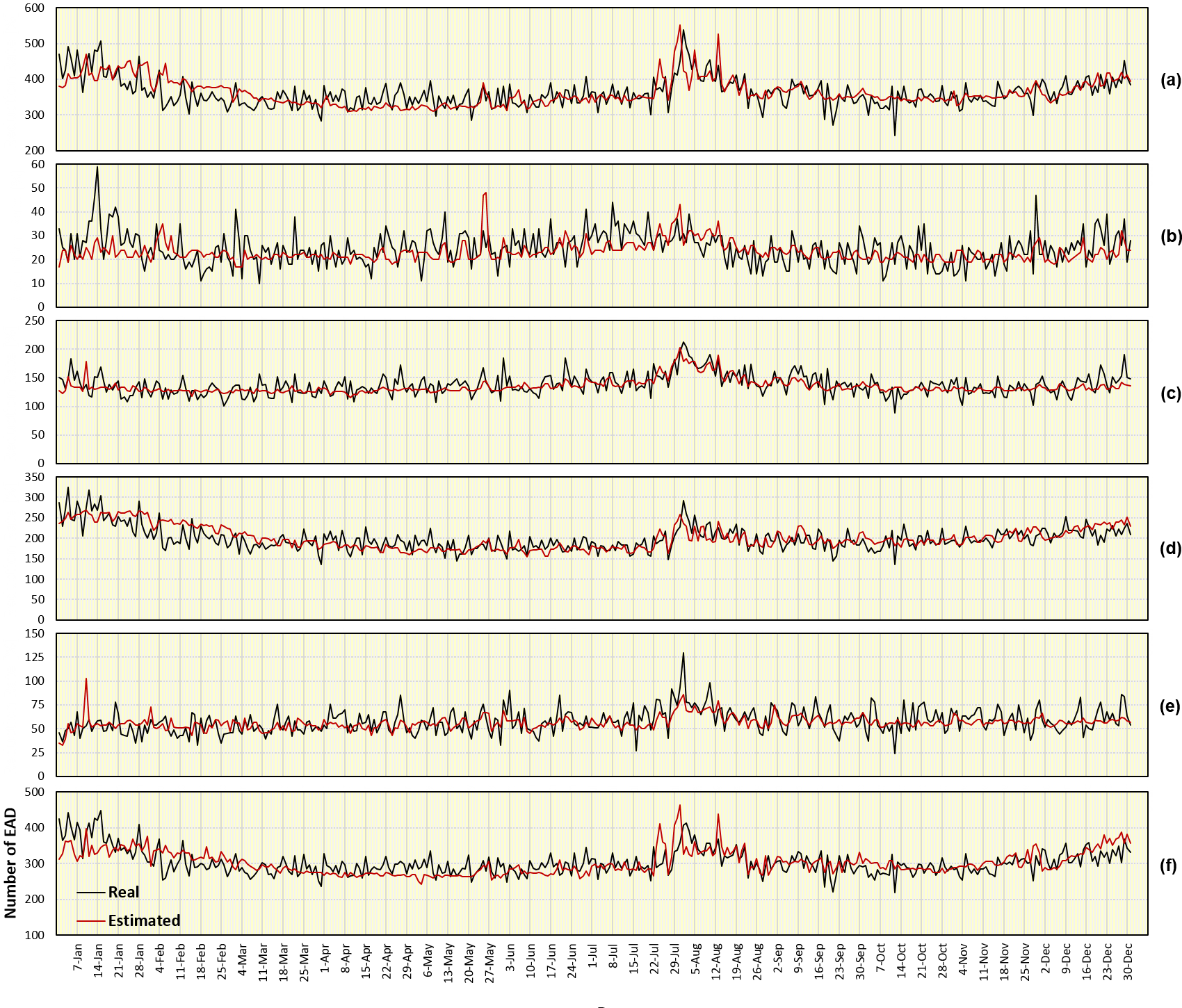}
\caption{Actual and estimated number of EADs during 2019 for (a) all groups and various age groups of (b) children, (c) adults, and (d) the elderly, as well as the location groups of (e) outdoor and (f) indoor patients.}
\label{2019LSTM}
\end{figure*}


\section{Results}

We investigated two scenarios that consider the pre- and ongoing pandemic periods. The pre-pandemic results are used to validate the feasibility of the LSTM architecture in estimating the number of daily EADs when considering the different categories and conditions. We further analyzed the effect of the Covid-19 pandemic and demonstrated a fine-tuning used to effectively handle bias factors resulting from such abnormalities. The forecast accuracy is validated using the correlation coefficient (CC) mean absolute error (MAE), which is defined as follows:

\begin{equation}
\textnormal{CC}(u,v)=\frac{n\sum_{i} u_i v_i - \sum_{i} u_i \sum_{i} v_i}{\sqrt{[n\sum_i u_i^2 -(\sum_i u_i)^2][n\sum_i v_i^2 -(\sum_i v_i)^2]}},
\end{equation}

\begin{equation}
\textnormal{MAE}(u,v)=\frac{1}{n} \sum_i \frac{|u_i-v_i|}{u_i}.
\end{equation}

Here, $u$ and $v$ are $n$-size vectors representing the real and estimated numbers of EADs, respectively.


\subsection{Daily EAD forecast during normal times (pre-pandemic)}

We consider data from April 1, 2014, to December 31, 2018, for training and 2019 data for testing. The output is split into different categories considering the pickup location (indoor/outdoor) and age categories: children (0 $\leq$ age $\leq$ 15), adult (15 $<$ age $\leq$ 65), and elderly (age $>$ 65). Training is considered through 500 epochs with a batch size of 8 samples. The results are presented in Fig.~\ref{2019LSTM}, and the descriptive statistics and quality metrics are listed in Table~\ref{Tab01}. It can be observed that elderly patients are estimated to have the highest accuracy compared with the other age categories. This may reflect the high sensitivity of elderly citizens to changes in environmental factors. In addition, the indoor patients can be estimated with a higher accuracy compared with outdoor patients because outdoor EADs include many cases that are unrelated to weather data such as road accidents.


\begin{table*}
\centering
\footnotesize
\caption{Descriptive statistics and error metrics for different data groups estimated for year 2019.}
\label{Tab01}
\setlength{\tabcolsep}{3pt}
\begin{tabular}{|c|c|c|c|c|c|c|c|c|c|c|c|c|c|c|}
\hline
\multicolumn{2}{|c|}{Group}  &Mean & Std. Error & Median & Mode & StDev & Kurtosis & Skewness & Range & Min & Max & Sum & CC & MAE\\
\hline\hline
\multirow{2}{*}{All} & {\bf Real} &361.06 & 2.08 & 356 & 364 & 39.74 & 2.14 & 1.03 & 294 & 243 & 537 & 131,788 & \multirow{2}{*}{0.591}&\multirow{2}{*}{0.071}\\
\cline{2-13}
    & {\bf Est.} &362.21 & 1.92 & 351 & 350 & 36.71 & 3.13 & 1.42 & 241 & 310 & 551 & 132,205 &&\\
\hline\hline

\multirow{2}{*}{Children} & {\bf Real} &24.58 & 0.36 & 24 & 27 & 6.96 & 1.34 & 0.65 & 49 & 10 & 59 & 8,973  & \multirow{2}{*}{0.284}&\multirow{2}{*}{0.232}\\
\cline{2-13}
    & {\bf Est.} &23.22 & 0.21 & 22 & 21 & 4.05 & 8.53 & 2.26 & 31 & 17 & 48 & 8,475  &&\\
\hline\hline

\multirow{2}{*}{Adult} & {\bf Real} & 137.57 & 0.97 & 135 & 124 & 18.51 & 1.14 & 0.77 & 124 & 89 & 213 & 50,213 & \multirow{2}{*}{0.619}&\multirow{2}{*}{0.083}\\
\cline{2-13}
    & {\bf Est.} & 134.89 & 0.64 & 131 & 127 & 12.14 & 6.92 & 2.39 & 88 & 115 & 203 & 49,235  &&\\
\hline\hline

\multirow{2}{*}{Elderly} & {\bf Real} & 198.91 & 1.50 & 195 & 184 & 28.61 & 2.49 & 1.16 & 189 & 135 & 324 & 72,602  & \multirow{2}{*}{0.672}&\multirow{2}{*}{0.091}\\
\cline{2-13}
    & {\bf Est.} & 200.87 & 1.44 & 195 & 197 & 27.56 & -0.43 & 0.69 & 115 & 154 & 269 & 73,316  &&\\
\hline\hline

\multirow{2}{*}{Outdoor} & {\bf Real} & 57.83 & 0.66 & 57 & 57 & 12.69 & 2.43 & 0.80 & 106 & 24 & 130 & 21,108 & \multirow{2}{*}{0.482}&\multirow{2}{*}{0.155}\\
\cline{2-13}
    & {\bf Est.} & 56.34 & 0.35 & 56 & 55 & 6.63 & 8.29 & 1.41 & 70 & 33 & 103 & 20,565 &&\\
\hline\hline

\multirow{2}{*}{Indoor} & {\bf Real}  & 303.23 & 1.89 & 297 & 301 & 36.02 & 2.11 & 1.18 & 228 & 219 & 447 & 110,680  & \multirow{2}{*}{0.526}&\multirow{2}{*}{0.083}\\
\cline{2-13}
    & {\bf Est.} & 302.50 & 1.74 & 295 & 275 & 33.17 & 2.14 & 1.21 & 221 & 243 & 464 & 110,413 &&\\
\hline
\end{tabular}
\end{table*}

\begin{figure*}
\centering
\includegraphics[width=\textwidth]{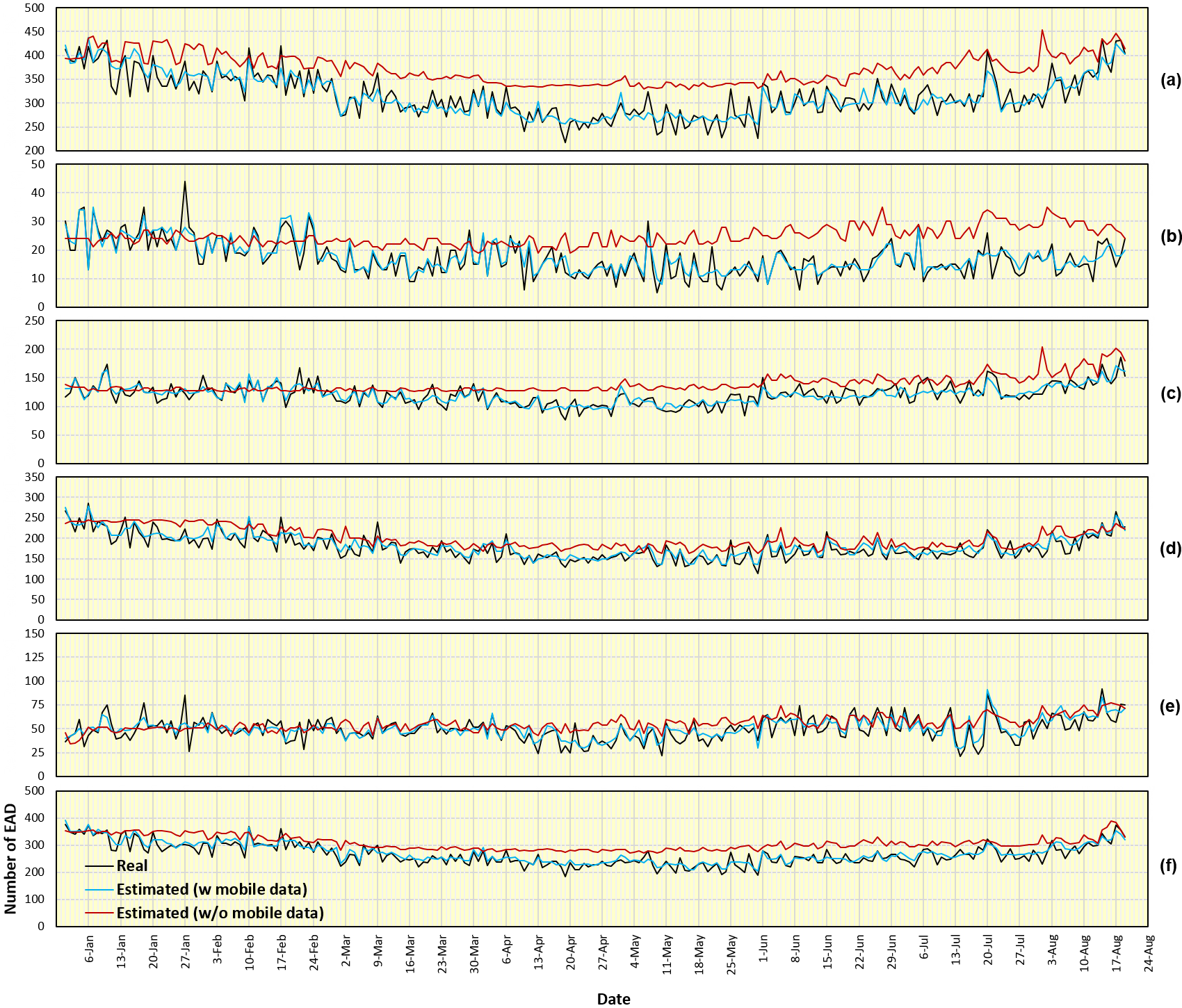}
\caption{Actual and estimated numbers of EADs in 2020 (January 1 to August 19) for (a) all groups and various age groups of (b) children, (c) adults, and (d) the elderly, as well as location groups of (e) outdoor and (f) indoor patients. Estimations that include mobile usage data provide more accurate results.}
\label{2020LSTM}
\end{figure*}

\begin{figure*}
\centering
\includegraphics[width=\textwidth]{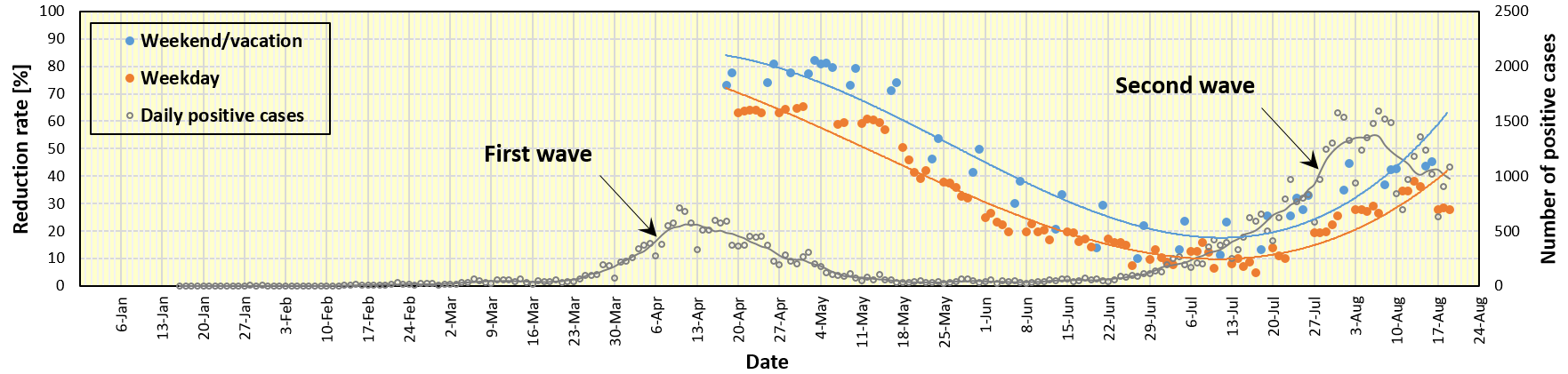}
\caption{Reduction rate of mobile users around Nagoya main station in year 2020 compared with the values measured in the previous year. A reduction in usage shows a consistent regression lines with the data split into working and non-working days. A reference curve in gray color demonstrate the number of daily positive COVID-19 cases during first and second waves in Japan (line represents a 7-days average).}
\label{ntt}
\end{figure*}


\subsection{Daily EAD forecast during the pandemic}

\begin{figure*}
\centering
\includegraphics[width=\textwidth]{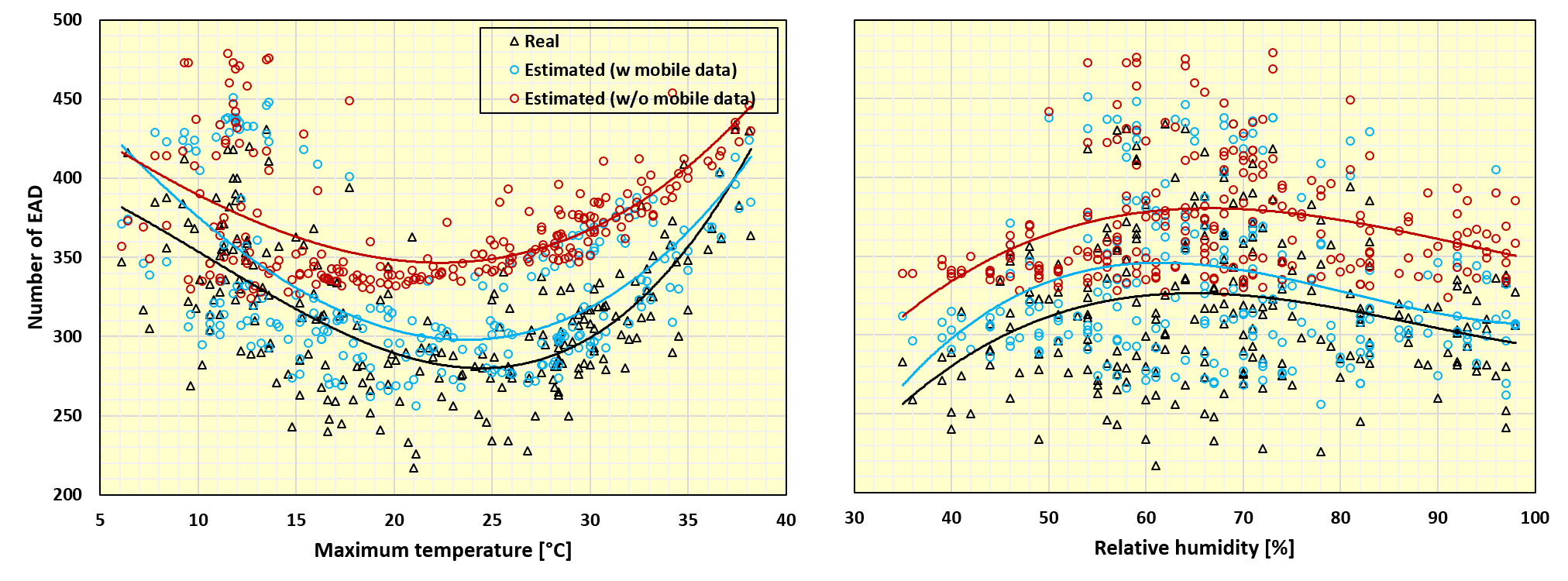}
\caption{Actual and estimated number of daily EADs (January 1 to August 19, 2020) associated with maximum temperature (left) and average absolute humidity (right). Curves represent the best-fitting polynomial (3$^\textnormal{rd}$ degree). Both weather factors, including mobile usage data, improve the forecast accuracy.}
\label{LSTMtmphum}
\end{figure*}

Under this scenario, we consider the data from April 1, 2014, to December 31, 2019, as the training set and the data from January 1 to August 19, 2020, for testing. The network is trained similarly to the first scenario setup; however, a significant error is found as the number of EADs is significantly decreased in 2020 (Fig.~\ref{2020LSTM}). To overcome this problem, the data on the mobile users’ location are included as an additional input term. The available data provided by NTT around Nagoya’s main station (April 18 to August 18, 2020) are shown in Fig.~\ref{ntt} along with profile of COVID-19 positive cases in Japan\footnote{\href{https://www.mhlw.go.jp/stf/covid-19/open-data.html}{https://www.mhlw.go.jp (21 Aug. 2020)}}. We assume that the mobile phone usage is 100\% during January 2020 where other missing data were linearly interpolated. Figure~\ref{2020LSTM} shows the forecast results with and without mobile data for different data groups. We observed a significant improvement in the estimation accuracy. Table~\ref{Tab02} presents the descriptive statistics and assessment metrics. These data clearly demonstrate that mobile data usage as a surrogate of social activities is useful in improving the forecasting of EADs in different data groups. To clearly demonstrate the forecast accuracy, a plot representing the estimated number of daily EADs in association with the daily maximum temperature and average relative humidity is shown in Fig.~\ref{LSTMtmphum}. Estimated data are demonstrated with almost the same pattern in both cases; however, including the mobile users’ location data significantly reduces the error caused by the abnormalities.

\begin{table*}
\centering
\footnotesize
\caption{Descriptive statistics and error metrics for different data groups estimated with ({\bf Est.+}) and without ({\bf Est.}) mobile data usage for year 2020.}
\label{Tab02}
\setlength{\tabcolsep}{3pt}
\begin{tabular}{|c|c|c|c|c|c|c|c|c|c|c|c|c|c|c|}
\hline
\multicolumn{2}{|c|}{Group}  &Mean & Std. Error & Median & Mode & StDev & Kurtosis & Skewness & Range & Min & Max & Sum & CC & MAE\\
\hline\hline
\multirow{3}{*}{All} & {\bf Real} & 315.52&	2.98&	312	&317	&45.42&-0.14&	0.48	&217&	217&	434&	73,200& -- & --\\
\cline{2-15}
& {\bf Est.} & 370.99	&1.98	&366	&337	&30.09	&-0.63	&0.57	&126	&328	&454&	86,069	&0.770	&0.191\\
\cline{2-15}
& {\bf Est.+} & 315.84	&2.81&	304	&312	&42.81	&-0.45	&0.65&	182	&255	&437	&73,276	&0.910	&0.047\\
\hline\hline

\multirow{3}{*}{Children} & {\bf Real} & 17.17&	0.42&	16	&15	&6.37	&0.94	&0.84	&39	&5	&44	&3,984& -- & --\\
\cline{2-15}
& {\bf Est.}  & 24.71	&0.21	&24	&24	&3.20	&0.38&	0.82	&16	&19	&35&	5,733&	0.003	&0.698\\
\cline{2-15}
& {\bf Est.+} & 17.66	&0.36&	16	&13	&5.49	&0.43&	0.94&	27&	8	&35	&4,097&	0.905&	0.140\\
\hline\hline

\multirow{3}{*}{Adult} & {\bf Real} & 120.56&	1.19	&120&	121	&18.12	&0.52	&0.46&	110&	76	&186	&27,970 & -- & --\\
\cline{2-15}
& {\bf Est.}  & 138.18&	0.98	&132	&127	&14.87	&4.76	&2.08	&79	&125	&204&	32,057	&0.429&	0.187\\
\cline{2-15}
& {\bf Est.+} & 120.59&	0.98&	120	&117&	14.97	&0.38	&0.54	&76	&95	&171	&27,976&	0.844	&0.066\\
\hline\hline

\multirow{3}{*}{Elderly} & {\bf Real} & 177.72	&1.99	&171	&166	&30.29&	0.53&	0.85&	172	&114	&286	&41,232  & -- & --\\
\cline{2-15}
& {\bf Est.}  & 199.33&	1.62	&190	&180	&24.65	&-1.01	&0.61&	82&	163&	245&	46,244&	0.774	&0.148\\
\cline{2-15}
& {\bf Est.+} & 182.11&	1.78	&177&	181&	27.18&	0.51	&0.76	&143	&135	&278	&42,250&	0.895&	0.065\\
\hline\hline

\multirow{3}{*}{Outdoor} & {\bf Real} & 49.46&	0.81&	49&	60&	12.30	&0.48	&0.33&	71	&21&	92&	11,474 & -- & --\\
\cline{2-15}
& {\bf Est.}  & 55.15&	0.49	&54	&51	&7.53&	0.51&	0.54	&43	&34	&77	&12,794	&0.446	&0.244 \\
\cline{2-15}
& {\bf Est.+} & 50.66	&0.64	&50	&49	&9.77	&1.06	&0.45	&62	&29&	91	&11,754	&0.812	&0.126\\
\hline\hline

\multirow{3}{*}{Indoor} & {\bf Real} & 266.06	&2.65	&261	&265	&40.38&	-0.02	&0.61	&191&	185	&376	&61,726 & -- & --\\
\cline{2-15}
& {\bf Est.}  & 307.67	&1.68	&302	&285&	25.65	&-0.34	&0.74	&119	&271	&390	&71,380	&0.796	&0.175\\
\cline{2-15}
& {\bf Est.+} & 271.90	&2.41	&265	&251&	36.64	&-0.06&	0.69	&186	&205&	391&	63,081	&0.918&	0.052 \\
\hline
\end{tabular}
\end{table*}

\subsection{Contribution of different variables}
\begin{figure*}
\centering
\includegraphics[width=\textwidth]{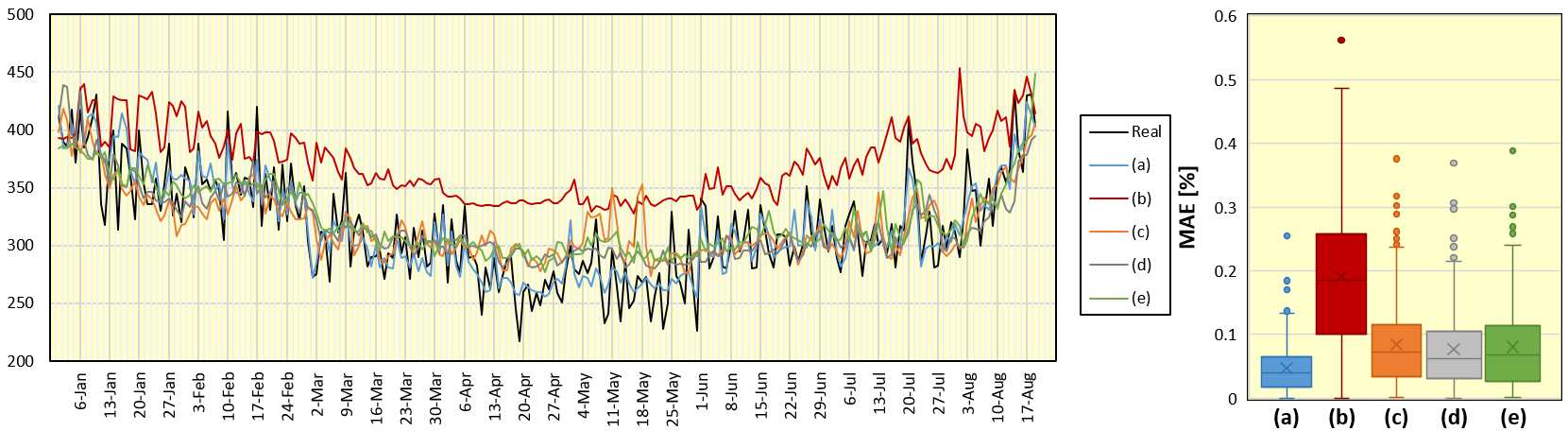}
\caption{Left is the actual and estimated number of daily EADs using (a) all data and when exclude (b) mobile usage, (c) temperature, (d) humidity and (e) day label data. Right is the box plots demonstrate MAE for cases of (a)-(e).}
\label{exvar}
\end{figure*}

To validate how different variables contribute to the estimation of daily EADs, we conducted an ablation study. The experimental data presented in Fig.~\ref{2020LSTM} is repeated with exclusion of single variable each. We consider exclusion of mobile usage data, maximum temperature, average humidity and day label (working day/off day). The network is retrained with these different set of variables and data of 2020 is estimated for each case. Results are shown in Fig.~\ref{exvar} along with box plots demonstrate MAE in each case. These results demonstrate that training using all variables leads to MAE of 4.70\% and mobility data that represent a surrogate of social activities during the pandemic is the most dominant variable that increase the MAE to 19.12\% when excluded. Excluding temperature, humidity and day label are of comparable importance and lead to increase MAE to values of 8.44\%, 7.66\% and 8.04\%, respectively.

\subsection{Long-term forecasting}
\begin{figure*}
\centering
\includegraphics[width=\textwidth]{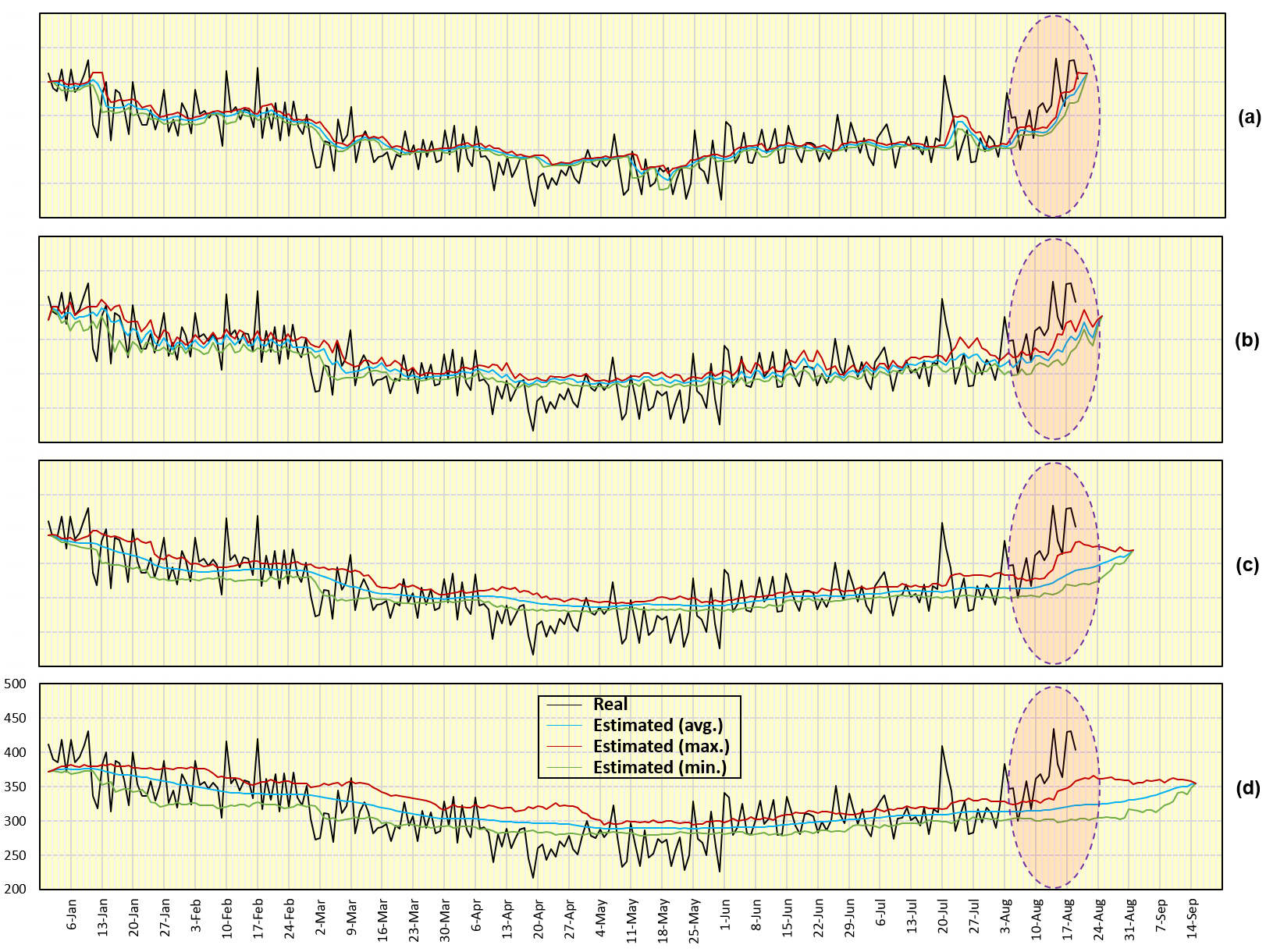}
\caption{Actual and estimated number of daily EADs in 2020 with forecasting of 3, 7, 14 and 28 days in (a)-(d), respectively. Average, maximum and minimum estimated values are used to demonstrate error range. Highlighted region with dashed ellipse clearly demonstrate that estimation error increase with long forecasting time period.}
\label{diffdays}
\end{figure*}

In some cases, it is required to have a long-term forecasting that demonstrate data beyond just a single day. The network architecture is designed to express estimation of $K$ successive days. The training session is repeated with different architectures with $K=3, 7, 14$ and $28$  with all other parameters fixed as those shown in Fig.~\ref{2020LSTM} and the number of EADs for 2020 is estimated once more. Figure~\ref{diffdays} demonstrate results obtained from different value of $K$. The future $K$ days are computed for each day from the beginning of 2020. Therefore, each day is presented with different values (1 to $K$). We demonstrate the average, minimum and maximum estimated daily values. It is clear from these data, that a good estimation can be achieved within small time period (e.g. 3 days), however, estimation error accumulates when the estimation period extended further. This is clear from the regions labeled with the dashed line ellipse in Fig.~\ref{diffdays}. MAE associated with $K=3, 7, 14$ and $28$ is 7.32\%, 7.79\%, 8.02\% and 8.46\%, respectively.


\section{Discussion}

In this study, the numbers of people transported by ambulance and based on population activities were evaluated for the planning of the third wave of COIVD-19 and future pandemics. As one notable feature in Japan, the government did not lock down the city, but requested people to apply voluntary constraints. Nagoya is a primal city in the third largest area, following the Kanto (Tokyo) and Kansai (Osaka) areas, in Japan. In addition, ambulance use is free in Japan; thus, the number of EADs corresponds approximately to the real number of patients who need such transport. 

The number of patients transported by ambulance during the state of the emergency was generally smaller than that during previous years. This difference may be attributable to the significant reduction in the activities of adults; the percentage of teleworking in Japan is normally approximately 1\%, whereas it was 30\% in April 2020. As a discrepancy in the activities of the population, teleworking became common around the city center but not in the suburbs. 

Patients younger than 65 years in age should be well correlated with the activities of the population in the city center. Similarly, a reduction was observed even in the elderly. We then demonstrated that environmental factors such as the maximum temperature and relative humidity can be used to estimate the number of people transported by ambulance. During a pandemic, special care (e.g., disinfection) is required even for emergency services. During this particular pandemic, ambulances were disinfected when transporting potential COVID-19 patients, at least in Nagoya. Thus, resource management was critical during the pandemic, and maintaining the number of dispatches below a certain level has been essential. During the pandemic, the number of patients transported was reduced by 20\% at maximum, whereas the amount of human activity around the central station was suppressed by 80\%. The findings here will be useful to estimate or plan ambulance allocations. The second SoE in Japan was declared active from January 7 up to March 7, 2021 (tentative schedule). The average daily mobility reduction rate in the first and second SoEs\footnote{data computed from January 8 to February 14, 2021} and the normal period in between at Nagoya main station was 63\%, 29.1\% and 18.8\%, respectively. This indicates the positive response of public (in different scales) during emergency calls to voluntarily reduce the social activities.

As a limitation of this study, the EAD data were not classified into specific diseases, some of which may be highly related to environmental factors (such as heat stroke or respiratory system failure), whereas others may not be related at the same level. However, splitting the data and applying a further generalization of the proposed model to handle this problem may remain as a future study. Moreover, a comparison with data obtained from different cities would provide a better understanding of the outlined framework.

The source code used in this study including trained network will be provided publicly after publication.


\section{Conclusion}

This study investigated the correlation between environmental factors such as ambient temperature, absolute humidity, and the daily number of EADs in Nagoya City, Japan. Data collected from April 2014 indicate a good correlation that may be potentially useful in future forecasting of required AED facilities based on weather data. A machine learning framework based on an LSTM network architecture was used for time-sampled forecasting, and interesting results were shown within a normal state. This finding presents for the first time an affordable method for estimating the number of EADs with environmental factors using the LSTM architecture. Moreover, a strong bias was recognized when forecasting the number of EADs required during the COVID-19 pandemic. To handle this problem, additional data indicating a reduction in mobile phone usage in major crowded areas, such as train stations, were used as a surrogate for the reduction of social activity during the pandemic. Including these data can significantly reduce the forecasting error during a time of uncertainty, such as during unexpected pandemics.

\bibliography{Refs1}
\end{document}